\documentclass{article}

\usepackage[final]{corl_2017} 

\usepackage{amsmath}
\usepackage{amssymb}  

\usepackage{booktabs}
\usepackage{graphicx}

\newcommand{\vect}[1]{\mathbf{ #1}}

\newcommand{\vectg}[1]{{\boldsymbol{ #1}}}

\newcommand{\ggo}{\ensuremath{\mathrm{g^2o}} }

\newcommand{\R}{\mathbb{R}}

\newcommand{\T}{^\mathsf{T}}

\newcommand{\argmin}{\operatornamewithlimits{argmin}}

\newcommand{\argmax}{\operatornamewithlimits{argmax}}

\newcommand{\RMSE}{\operatorname{RMSE}}
\newcommand{\RMSEpos}{\operatorname{RMSE}_\text{pos}}

\newcommand{\vA}{\vect{A}}

\newcommand{\vC}{\vect{C}}
\newcommand{\vD}{\vect{D}}

\newcommand{\vK}{\vect{K}}

\newcommand{\vP}{\vect{P}}
\newcommand{\vQ}{\vect{Q}}
\newcommand{\vR}{\vect{R}}

\newcommand{\vT}{\vect{T}}
\newcommand{\vU}{\vect{U}}
\newcommand{\vV}{\vect{V}}

\newcommand{\ve}{\vect{e}}

\newcommand{\vl}{\vect{l}}

\newcommand{\vq}{\vect{q}}

\newcommand{\vt}{\vect{t}}
\newcommand{\vu}{\vect{u}}

\newcommand{\vw}{\vect{w}}
\newcommand{\vx}{\vect{x}}

\newcommand{\vz}{\vect{z}}

\newcommand{\vpi}{\vectg{\pi}}

\newcommand{\vSigma}{\vectg{\Sigma}}

\newcommand{\cD}{\mathcal{D}}

\newcommand{\cL}{\mathcal{L}}

\newcommand{\cN}{\mathcal{N}}

\title{Dual Quadrics
from Object Detection Bounding Boxes
as Landmark Representations in SLAM}

\author{
  Niko S\"underhauf and Michael Milford\\  
  ARC Centre of Excellence for Robotic Vision\\
  Queensland University of Technology (QUT), Brisbane, Australia\\
  \texttt{niko.suenderhauf@qut.edu.au} \\
}

\begin{document}
\maketitle

\begin{abstract}
Research in Simultaneous Localization And Mapping (SLAM) is increasingly moving towards richer world representations involving objects and high level features that enable a semantic model of the world for robots, potentially leading to a more meaningful set of robot-world interactions. Many of these advances are grounded in state-of-the-art computer vision techniques primarily developed in the context of image-based benchmark datasets, leaving several challenges to be addressed in adapting them for use in robotics. In this paper, we derive a formulation for Simultaneous Localization And Mapping (SLAM) that uses dual quadrics as 3D landmark representations, and show how 2D bounding boxes (such as those typically obtained from visual object detection systems) can directly constrain the quadric parameters. Our paper demonstrates how to jointly estimate the robot pose and dual quadric parameters in factor graph based SLAM with a general perspective camera, and covers the use-cases of a robot moving with a monocular camera with and without the availability of additional depth information.
\end{abstract}


\section{Introduction}

In recent years, impressive vision-based object detection performance improvements have resulted from the "rebirth" of Convolutional Neural Networks (ConvNets). Building on the seminal work by Krizhevsky et al.~\citep{Krizhevsky12} and earlier work~\citep{LeCun98, LeCun04}, several other groups (e.g. \citep{Sermanet13, Girshick14, Szegedy15, Ren15, Liu16, He17}) have increased the quality of ConvNet-based methods for object detection. Recent approaches have even reached human performance on the standardized ImageNet ILSVRC benchmark \citep{Russakovsky15} and continue to push the performance boundaries on other benchmarks such as COCO~\citep{Lin14}.

Despite these impressive developments, the Simultaneous Localization And Mapping community (SLAM) has not yet fully adopted the newly arisen opportunities to create semantically meaningful maps. SLAM maps typically represent \emph{geometric} information, but do not carry immediate object-level \emph{semantic} information. Semantically-enriched SLAM systems are appealing because they increase the richness with which a robot can understand the world around it, and consequently the range and sophistication of interactions that that robot may have with the world, a critical requirement for their eventual widespread deployment at work and in homes.

Semantically meaningful maps should be object-oriented, with objects as the central entities of the map. \emph{Quadrics}, i.e. 3D surfaces such as ellipsoids, are ideal landmark representations for object-oriented semantic maps. 
In contrast to more complex object representations such as truncated signed distance fields~\citep{Curless96}, quadrics have a very compact representation and can be manipulated efficiently within the framework of projective geometry. Quadrics also capture information about the size, position, and orientation of an object, and can serve as anchors for more detailed 3D reconstructions if necessary.
They are also appealing from an integration perspective: in their \emph{dual} form, quadrics can be constructed directly from object detection bounding boxes and conveniently incorporated into a factor graph based SLAM formulation.

In this paper we make the following contributions. We first show how to parametrize object landmarks in SLAM as \emph{dual quadrics}. We then demonstrate that the bounding boxes created by visual object detection systems such as Faster R-CNN~\citep{Ren15}, SSD~\citep{Liu16}, or Mask R-CNN~\citep{He17}, can directly constrain the dual quadric parameters. To incorporate quadrics into SLAM, we derive a factor graph-based SLAM formulation that jointly estimates the dual quadric and robot pose parameters. Our evaluation shows how object detections and the dual quadric parametrization aid the SLAM solution, both in the monocular case and when depth measurements are available.

Previous work \citep{Rubino17} utilized dual quadrics as a parametrization for landmark \emph{mapping}\footnote{referred to as 'landmark localisation' in \citep{Rubino17}} only, or was limited to an orthographic camera \citep{Crocco16}: in this new work we perform full SLAM, i.e. Simultaneous Localization \emph{And} Mapping, with a general  \emph{perspective} camera. We are consequently able to exploit dual quadric landmarks to perform loop closures and correct the accumulated odometry error in the estimated camera trajectory.  Furthermore, previous work \citep{Rubino17, Crocco16} required ellipse fitting as a pre-processing step: here we show that dual quadrics can be estimated in SLAM directly from bounding boxes.

\section{Related Work}
\paragraph{Maps and Landmark Representations in SLAM}
Most current SLAM systems represent the environment as a collection of distinct geometric points that are used as landmarks. ORB-SLAM~\citep{Mur15, Mur16} is one of the most prominent recent examples for such a point-based visual SLAM system. Even \emph{direct} visual SLAM approaches~\citep{Engel14, Whelan15} produce point cloud maps, albeit much denser  than previous approaches. Other authors explored the utility of higher order geometric featuers such as line segments \citep{Lemaire07b} or planes \citep{Kaess15}.

A commonality of all those geometry-based SLAM systems is that their maps carry \emph{geometric} but no immediate \emph{semantic} information. 
An exception is the seminal work by Salas-Moreno et al.~\citep{Salas13}. This work proposed a truly object oriented SLAM system by using real-world \emph{objects} such as chairs and tables as landmarks instead of geometric primitives. \citep{Salas13} detected these objects in RGB-D data by matching 3D models of known object classes. In contrast to \citep{Salas13}, the approach presented in this paper does not require a-priori known object CAD models, but instead uses general purpose visual object detection systems, typically based on deep convolutional networks, such as \citep{Liu16, Redmon16, Ren15}.

SemanticFusion~\citep{McCormac16} recently demonstrated how a dense 3D reconstruction obtained by SLAM can be enriched with semantic information. This work, and other similar papers such as \citep{Pham15} add semantics to the map \emph{after} it has been created. The maps are not object-centric, but rather dense point clouds, where every point carries a semantic label, or a distribution over labels. In contrast, our approach uses objects as landmarks inside the SLAM system, and the resulting map consists of objects encoded as quadrics.

\paragraph{Dual Quadrics as Landmark Representations}
The connection between object detections and dual quadrics was recently investigated by \citep{Crocco16} and ~\citep{Rubino17}, which are most directly related to our work.
Crocco et al.~\citep{Crocco16} presented an approach for estimating dual quadric parameters from object detections in closed form. Their method however is limited to orthographic cameras, while our approach works with perspective cameras, and is therefore more general and applicable to robotics scenarios. Furthermore, \citep{Crocco16} requires an ellipse-fitting step around each detected object. In contrast, our method can estimate camera pose and quadric parameters directly from the bounding boxes typically produced by object detection approaches such as \citep{Liu16, Redmon16, Ren15}.

As an extention of \citep{Crocco16}, Rubino et al.~\citep{Rubino17} described a closed-form approach to recover dual quadric parameters from object detections in multiple views. Their method can handle perspective cameras, but does not solve for camera pose parameters. It therefore performs only landmark \emph{mapping} given known camera poses. In contrast, our approach performs full Simultaneous Localization And Mapping, i.e. solving for camera pose and landmark pose and shape parameters simultaneously. Similar to \citep{Crocco16}, \citep{Rubino17} also requires fitting ellipses to bounding box detections first.

\section{Dual Quadrics as Landmarks in SLAM}
This section of our paper explains the connections between object detection bounding boxes and dual quadrics step by step. For a more in-depth coverage we refer the reader to textbooks on projective geometry such as \citep{Hartley04}.
\subsection{From Object Detections to Plane Envelopes in 3D}
Typical state-of-the-art object detection approaches generate a bounding box around the detected object. This bounding box can very easily be represented by its four corner points, as a set of homogeneous points $\cD = \{\vx_1, \vx_2, \vx_3, \vx_4\}$. We can also represent the bounding box by the four lines $\cL = \{\vl_1, \vl_2, \vl_3, \vl_4\}$ that are defined by connecting neighboring corner points. These lines can be parametrized by the cross products between two corner points, so that $\vl_1 = \vx_1 \times \vx_2$, $\vl_2 = \vx_2 \times \vx_3$, $\vl_3 = \vx_3 \times \vx_4$, and $\vl_4 = \vx_4 \times \vx_1$. 

Each of these four lines backprojects to a plane $\vpi_i$ in 3D $\vpi_i = \vP\T\vl_i$.
The camera projection matrix $\vP$ is defined as $\vP = \vK[\vR | \vt]$ and contains the camera intrinsic parameters $\vK$, and the camera pose given by $\vR$ and $\vt$. Since both $\vx_i$ above and $\vl_i$ are homogeneous 3-vectors, and $\vP$ is a $3\times 4$ matrix, the planes $\vpi_i$ are homogeneous 4-vectors.

We can see that a single observation of an object bounding box gives us four 3D planes that pass through the camera center and the sides of the bounding box on the image plane. The object is contained by the volume between these planes: i.e. the planes form an \emph{envelope} around the object.

The object's distance from the camera cannot be observed directly from a single observation, unless additional information is given. This is equivalent to the fact that a single observation of a point landmark gives rise to a single ray in 3D space, constraining the position of the point landmark to that ray, but with unknown distance. Similar to point landmarks, repeatedly observing the object from a number of sufficiently different viewpoints should allow us to retrieve more information about its full position\footnote{Alternatively, with an estimate of the size of the object in 3D, it is possible to estimate its distance from the camera, given a single object detection. The semantic knowledge contained in an object detection allows us to exploit prior knowledge in the form of a distribution over object sizes, given the semantic class of a detected object.}. 

\subsection{From Plane Envelopes to Dual Quadrics}
\label{sec:quadrics}
Quadrics are surfaces in 3D space that are defined by a $4 \times 4$ symmetric matrix $\vQ$, so that all points $\vx$ on the quadric fulfill $\vx\T\vQ\vx = 0$. Examples for quadrics are bodies such as spheres, ellipsoids, hyperboloids, cones, or cylinders. 

While the above definition of a quadric concentrates on the points $\vx$ on the quadric's surface, a quadric can also be defined by a set of tangential \emph{planes} such that the planes form an envelope around the quadric. This \emph{dual} quadric $\vQ^*$ is defined as 
\begin{equation}
\vpi\T\vQ^*\vpi = 0
\end{equation}
Every quadric $\vQ$ has a corresponding dual form $\vQ^* = \operatorname{adjoint}(\vQ)$, or $\vQ^* = \vQ^{-1}$ if $\vQ$ is invertible.

The $4\times 4$ matrix $\vQ^*$ is symmetric, and therefore has 10 degrees of freedom. However, since all equations are defined in homogeneous form, we can fix one degree of freedom (the overall scale factor), 
which results in a 9 degree of freedom representation $\vq = (q_1, q_2, ... , q_9)\T$ so that 
\begin{equation}
\label{eq:Q_from_q}
\vQ^*_{(\vq)} = 
\begin{pmatrix}
q_1 & q_2 & q_3 & q_4 \\
q_2 & q_5 & q_6 & q_7 \\
q_3 & q_6 & q_8 & q_9 \\
q_4 & q_7 & q_9 & 1 \\
\end{pmatrix}
\end{equation}

\subsection{From Dual Quadrics to Dual Conics and Back to Object Detections}

When a quadric is projected onto an image plane, it creates a dual \emph{conic}, following the simple rule 
\begin{equation}
\vC^* = \vP\vQ^*\vP\T
\label{eq:PQP}
\end{equation}
Again, $\vP = \vK[\vR | \vt]$ is the camera projection matrix that contains intrinsic and extrinsic camera parameters.
Conics are the 2D counterparts of quadrics and form shapes such as circles, ellipses, parabolas, or hyperbolas. Just like quadrics, they can be defined in a primal form via points  ($\vx\T\vC\vx=0$), or in dual form using tangent lines: 
\begin{equation}
\vl\T\vC^*\vl=0    
\label{eq:lCl}
\end{equation}
This dual form is of interest to us, since it expresses the fact that a dual conic is formed by a set of tangential lines, and all lines tangential to the conic fulfill the above equation. Combining (\ref{eq:PQP}) and (\ref{eq:lCl}) enables us to form an equation that constrains the dual quadric $\vQ^*$ 
given a line $\vl$:
\begin{equation}
\vl\T\vP\vQ^*\vP\T\vl = 0 
\end{equation}
Remembering that a single object detection bounding box gives rise to 4 image lines $\vl_{1..4}$, we understand that every object detection produces 4 such constraints on the dual quadric parameters. If we accumulate enough object detections, we can find a solution for the unknown parameters of the quadric $\vQ*$, if the camera parameters in $\vP$ are known. Since $\vQ^*$ has 9 degrees of freedom, we require at least 3 observations from different viewpoints to obtain a unique solution.

\subsection{Finding a Least Squares Solution for $\vQ^*$}

While the above equation would hold exactly only in a noise-free case where both the parameters of the line $\vl$ and the camera model are known perfectly, in general we will encounter noisy observations. Furthermore, in many situations there will be more than three observations available. In these cases, we can find the optimal dual quadric $\hat\vQ^*$ by solving a least squares optimization problem:
\begin{equation}
\hat{\vQ}^*_{(\vq)} = \argmin_{\vq}\|\vl_{ik}\T\vP_i\vQ^*_{(\vq)}\vP_i\T\vl_{ik}\|^2
\label{eq:Q_lsq}
\end{equation}
Here we write $\vl_{ik}$ to indicate the $k$-th line observed by the camera with camera matrix $\vP_i$. The lines are formed by the found object detection bounding boxes, $k=1..4$.  The term $\vl\T\vP\vQ^*\vP\T\vl$ is in essence a \emph{reprojection error}, and minimizing it leads us to the optimal quadric parameters.

This optimization problem can be solved if the camera matrices $\vP_i$ are known. By expanding $\vP_i$ into its components, we see that we can also formulate an optimization problem that solves for the dual quadric \emph{and} the extrinsic camera parameters jointly:
\begin{equation}
\hat{\vQ}^* ,\hat\vR_i, \hat\vt_i = \argmin_{\vq, \vR_i, \vt_i}\|\vl_{ik}\T\vK[\vR_i|\vt_i]\vQ^*_{(\vq)}[\vR_i|\vt_i]\T\vK\T\vl_{ik}\|^2
\end{equation}
However, by examining the number of unknown variables, we see that this is an underconstrained problem and cannot be solved unless additional information is available. We will investigate how we can utilize such additional information in the form of odometry measurements in a SLAM context in Section \ref{sec:quadric_SLAM}.

\section{SLAM with Dual Quadric Landmark Representations}
\label{sec:quadric_SLAM}

\subsection{General Problem Setup}
We will set up a SLAM problem where we have odometry measurements $\vu_i$ between two successive poses $\vx_i$ and $\vx_{i+1}$, so that $\vx_{i+1} = f(\vx_i, \vu_i) + \vw_i$.
Here $f$ is a usually nonlinear function that implements the motion model of
the robot and the $\vx_i$ and $\vx_j$ are the unknown robot poses. $\vw_i$ are zero-mean Gaussian error terms with covariances $\Sigma_i$.
The source of these odometry measurements $\vu_i$ is not of concern for the following discussion, and various sources such as wheel odometers or visual odometry are possible. Likewise, this general formulation covers the 2D or 3D case, as the robot poses $\vx_i$ can be parametrized as $\vx_i \in \text{SE}(2)$ or $\vx_i \in \text{SE}(3)$, with $f$ and $\vu_i$ defined accordingly.

We furthermore observe a set of image lines $L = \{\vl_{ijk}\}$ with $k=1..4$. We use this notation to indicate one of the four lines being observed from pose $\vx_i$, originating from a bounding box around an object $j$. Notice that we assume the problem of \emph{data association} is solved, i.e. we can identify which physical object $j$ the detection originates from\footnote{For a discussion of SLAM methods robust to data association errors see the relevant literature such as \citep{Suenderhauf12e, Agarwal13}. The methods discussed for pose graph SLAM can be adopted to the landmark SLAM considered here.}.

\subsection{Dual Quadric Parametrization for SLAM}

As discussed in Section \ref{sec:quadrics}, a dual quadric is a $4\times 4$ symmetric matrix. Because the overall scale factor, i.e. the last element $\vQ^*(4,4)$, can be fixed to an arbitrary value, the dual quadric has 9 degrees of freedom. We can therefore represent a dual quadric with a 9-vector $\vq$ and reconstruct the full dual quadric $\vQ^*$ as defined in equation (\ref{eq:Q_from_q}). Interestingly, the last column of $\vQ^*$ represents the quadric centroid as a homogeneous 4-vector. If we keep the scale factor of $\vQ^*$ fixed to 1 in all calculations, we can easily retrieve the quadric centroid from $\vq$ directly, as $(q_4, q_7, q_9)\T$.

\subsection{Building a Factor Graph Representation}

The conditional probability distribution over all robot poses $X=\{\vx_i\}$, and landmarks $Q=\{\vq_j\}$, given the observations $U=\{\vu_i\}$, and $L=\{\vl_{ijk}\}$ can be factored as
\begin{equation}
   P(X,Q|U,L) \propto \underbrace{\prod_i P(\vect{x}_{i+1} | \vect{x}_i,
   \vect{u}_{i})}_\text{Odometry Factors}
   \cdot
   \underbrace{\prod_{ijk} P(\vq_j | \vect{x}_i, \vect{l}_{ijk})}_\text{Landmark Factors}
  \label{eq:SLAM:posegraph:probability}
\end{equation}
This factored distribution can be conveniently modelled as a factor graph \cite{Kschischang01}. 

Given the sets of observations $U,L$ we seek the \emph{optimal}, i.e. maximum a posteriori (MAP) configuration of robot poses and dual quadrics, $X^*$, $Q^*$ to solve the landmark SLAM problem represented by the factor graph. This MAP variable configuration is equal to the mode of the joint probability distribution $P(X,Q)$. In simpler words, the MAP solution is the point where that distribution has its maximum.

\subsection{Finding the Maximum-a-Posteriori Solution}

We know from equation (\ref{eq:Q_lsq}) that the optimal solution to the dual quadric parameters $\vq_j$ is given by the solution to the least squares problem $\vq^*_j = \argmin_{\vq_j} \sum_{ik} \|\vl_{ijk}\T\vP_i\vQ^*_{(\vq_j)}\vP_i\T\vl_{ijk}\|^2$. Since $\vq^*_j$ is the optimal solution to the dual quadric parameters given the observtions $\vl_{ijk}$ and camera matrices $\vP_i$, we can postulate that $\vq^*_j$ also maximizes the probability distribution $P(\vq_j | \vx_i, \vl_{ijk})$:
\begin{equation}
   \vq^*_j = \argmin_{\vq_j}\|\vl_{ijk}\T\vP_{(\vx_i)}\vQ^*_{(\vq_j)}\vP_{(\vx_i)}\T\vl_{ijk}\|^2 = \argmax_{\vq_j} P(\vq_j | \vx_i, \vl_{ijk})
\end{equation}
with $\vP_{(\vx_i)}$ indicating the camera matrix according to the pose parameters in $\vx_i$, so that $\vP_{(\vx_i)} = \vK[\vR_i | \vt_i]$.

Making the common assumption that the odometry factors $P(\vx_{i+1} | \vx_i, \vu_i)$ are Gaussian, i.e. $\vx_{i+1} \sim \cN(f(\vx_i, \vu_i), \vSigma_i)$, the optimal variable configuration ${X^*, Q^*}$ can be determined by maximizing the joint probability from above: 
\begin{align}
  X^*, Q^* &=\argmax_{X,Q} P(X,Q|U,L) 
      =\argmin_{X,Q} -\log P(X,Q|U,L) \nonumber\\
      &=\argmin_{X,Q} 
      \underbrace{\sum_i \|f(\vect{x}_i, \vect{u}_i) \ominus \vect{x}_{i+1}\|^2_{\Sigma_{i}}}_\text{Odometry Factors}
      + \underbrace{\sum_{ijk} \|\vl_{ijk}\T\vP_{(\vx_i)}\vQ^*_{(\vq_j)}\vP_{(\vx_i)}\T\vl_{ijk}\|^2_{\Lambda_{ijk}}}_\text{Quadric Landmark Factors}
      \label{eq:SLAM_lsq}
\end{align}
This is a nonlinear least squares problem, since we seek the minimum over a sum of squared terms.

Here and throughout the paper $\|a-b\|^2_\Sigma$ denotes the squared Mahalanobis distance with covariance $\Sigma$. We use the $\ominus$ operator in the odometry factor to denote the difference operation is carried out in SE(2) or SE(3) space, not in $\R^2$ or $\R^3$.

Nonlinear least-squares problems such as (\ref{eq:SLAM_lsq}) can be solved using a variety of methods like
Levenberg-Marquardt, Gauss-Newton or Powell's Dog-Leg. These approaches
iteratively solve the problem by repeatedly linearizing it and updating the
current estimates of the unknown variables until convergence. At their heart, these methods
rely on a factorization (either QR or Cholesky) of the Jacobian associated with the factor graph.

Specialized solvers that exploit the sparse nature of the factorization (i.e. the sparse structure of the
Jacobians) can solve typical problems with thousands of variables very efficiently. Examples of convenient C++ frameworks 
that contain such solvers and can be easily applied to a number of different problem domains are GTSAM\footnote{GTSAM is probably the most widely adopted framework in the robotics community.}~\cite{gtsam} or \ggo \cite{Kummerle11}. 

\subsection{Factors for Observing the Relative Landmark Position}
The SLAM problem represented by equation (\ref{eq:SLAM_lsq}) uses odometry and bounding box measurements. If additional information about the position of the observed landmarks relative to the robot are available (e.g. from a depth camera or stereo camera), we can augment the problem with an additional factor to capture these measurements. We will denote these 
relative position measurements $\vz_{ij} \in \R^3$ and augment (\ref{eq:SLAM_lsq}) to the following optimization problem:
\begin{align}
  X^*, Q^* &=\argmax_{X,Q} P(X,Q|U,L,Z) 
      =\argmin_{X,Q} -\log P(X,Q|U,L,Z) \nonumber\\
      &=\argmin_{X,Q} 
      \underbrace{\sum_i \|\ve_i^\text{odo}\|^2_{\Sigma_{i}}}_\text{Odometry Factors}
      + \underbrace{\sum_{ijk} \|\ve_{ijk}^\text{quadric}\|^2_{\Lambda_{ijk}}}_\text{Quadric
      Landmark Factors} 
      + \underbrace{\sum_{ij} \|\vz_{ij} - \vT_i (\vq_j^\text{t})\|^2_{\Omega_{ij}}}_\text{Relative Position Factors}
      \label{eq:SLAM_lsq2}
\end{align}
With $\vq_j^\text{t} = (q_4, q_7, q_9)\T$ representing the estimated quadric centroid, and $\vT_i$ the transformation to transform the centroid coordinates (given in the world frame) into the local robot frame, as estimated by $\vx_i$.

\subsection{Variable Initialization}
All variable parameters $\vx_i$ and $\vq_j$ must be initialized in order for the incremental solvers to work. While the robot poses $\vx_i$ can be initialized to an initial guess obtained from the raw odometry measurements $\vu_i$, initializing the dual quadric landmarks $\vq_j$ requires more consideration. 

It is possible to initialize $\vq_j$ with the least squares fit to its defining equation:
\begin{equation}
  \label{eq:q_init_1}
  \vpi_{ijk}\T\vQ^*_{(\vq_j)}\vpi_{ijk} = 0  
\end{equation}
First we can form the homogeneous vectors defining the planes $\vpi_{ijk}$ using the landmark bounding box observations and resulting lines $\vl_{ik}$, by projecting them according to $\vpi_{ijk} = \vP_i\T\vl_{ijk}$. Here the camera matrix $\vP_i$ is formed using the initial camera pose estimates $\vx_i$ obtained from the odometry measurements. Exploiting the fact that $\vQ^*_{(\vq_j)}$ is symmetric, we can rewrite (\ref{eq:q_init_1}) for a specific $\vpi_{ijk}$ as:
\begin{equation}
  \label{eq:q_init_2}
  (\pi_1^2, \pi_1\pi_2, \pi_1\pi_3, \pi_1,  \pi_2^2, \pi_2\pi_3, \pi_2, \pi_3^2, \pi_3, \pi_4) \cdot (q_1, q_2, ..., q_{10})\T = 0  
\end{equation}
By collecting all these equations that originate from multiple views $i$ and planes $k$, we obtain a linear system of the form $\vA_j \vq_j = 0$ with $\vA_j$ containing the coefficients of all $\vpi_{ijk}$ associated with observations of landmark $\vq_j$ as in (\ref{eq:q_init_2}). A least squares solution $\hat\vq_j$ that minimizes $\|\vA_j \vq_j\|$ can be obtained as the last column of $\vV$, where $\vA_j \vq_j = \vU\vD\vV\T$ is the SVD of $\vA_j \vq_j$.

While this approach should in general give the optimal initialization point, we found it tends to result in degenerate solutions in the case of close-to-planar camera trajectories. This effect was also mentioned in \citep{Rubino17}. Since our evaluation relies on planar trajectories (as explained in the next section), the degenerate initializations for $\vq_i$ led to highly unstable convergence behaviour when solving the overall SLAM system, often creating solutions inferior to the initialization. We therefore initialized all $\vq_j$ as $(1, 0, 0, 0, 1, 0, 0, 1, 0)\T$, which results in an identity matrix for $\vQ^*_i$.

\section{Experiments and Evaluation}

\subsection{Synthetic Dataset Description}
We created a synthetic dataset, simulating a robot with a camera driving a 130 meter long trajectory in an environment with 10 randomly placed object landmarks. While the robot's movements are constrained to the ground plane $z=0$, so that $\vx_i \in \text{SE(2)}$, the landmarks are placed in 3D with the $z$ component chosen from a Normal distribution with standard deviation 0.3. The landmarks are cubes, with side lengths drawn from a Normal distribution according to $\max(0.2, \cN(0.5, 0.3))$. The camera was simulated with a focal length of 15 mm, a pixel size of 10e-6 m, and a resolution of $1280\times 1024$. If the projected objects were larger then 100 pixels in the camera image, we assumed it to be detected by a ConvNet-based object detector such as \citep{Ren15,Liu16} and generated a bounding box observation. The corners of the bounding box were corrupted by Gaussian noise with standard deviation of 1 pixel.

The robot was driven in two loops, with the camera facing 90 degrees sideways to the left. The odometry measurements $(v, \omega)\T$ at each timestep were corrupted by Gaussian noise, with standard deviations of 0.02 respectively. During the turns, $\omega$ was corrupted by noise with $\sigma=0.1$. Relative position measurements $\vz_{ij}$ were also corrupted with Gaussian noise, with a relatively large standard deviation of 10 cm to account for the fact that in reality a depth sensor would measure various parts on the surface of an irregular shaped object, but could not observe the object centroid directly.

For the evaluation, we repeatedly generated noisy trajectories and landmark observations, for a total of 50 trials. 
Figure \ref{fig:maps} shows some sample trajectories, illustrating the range of variance introduced by the odometry errors, and the optimized trajectories and landmark positions obtained from the SLAM solution.

\subsection{Evaluation Metrics}
\paragraph{Trajectory Quality} To evaluate the quality of the estimated robot trajectory, we calculate the average deviation of every estimated robot position from its ground truth. This is identical to the root mean squared error $\RMSEpos = \frac{1}{n}\sum_{i=1}^n \sqrt{\|\vx_i^{x,y} - \hat{\vx}_i^{x,y}\|^2}$ where $\vx^{x,y}_i = (x_i, y_i)\T$ is the estimated robot position and $\hat{\vx}{}^{x,y}_i$ is the respective ground truth position.

\paragraph{Landmark Quality} We evaluate the estimated landmark parameters by comparing the distance of the estimated landmark \emph{centroids} from its true position. The error in the estimated quadric centroid position is calculated as 
$\RMSE_\text{LM} = \frac{1}{n}\sum_{i=1}^n \sqrt{\|\vq_j^\text{t} - \hat{\vq}_j^\text{t}\|^2}$ where 
$\vq_j^\text{t} = (q_4, q_7, q_9)\T$ encodes the centroid of the estimated quadric $\vq_j$ as discussed earlier, and $\hat{\vq}_j^\text{t}$ is the ground truth centroid position.

We furthermore compare the volumes spanned by the estimated dual quadric with the ground truth volume of the observed landmarks.
Since the ground truth landmarks are cubes, a shape that cannot be represented exactly by quadrics, this measure can only give an approximation of the quality of the fit. The same would be true for the intersection over union, for the same reasons. The closest fit is to form a cube with a side length equal to the smallest value in the quadric's diagonal shape matrix $\vD$ after decomposing $\vQ^*_{(\vq_j)} = \vU\T\vD\vU$. This corresponds to the smallest cube that can be fully contained in the quadric.

\subsection{Results and Discussion}

\begin{table}[bt]
    \centering
    \begin{tabular}{@{}lccccccccc@{}}
        \toprule
        \multicolumn{3}{r}{\bf Initial (Odometry-only)}  & \hspace{0.2cm} & \multicolumn{6}{c}{\bf SLAM Solution} \\
        & \multicolumn{2}{c}{$\RMSEpos$} &  & \multicolumn{2}{c}{$\RMSEpos$} & \multicolumn{2}{c}{$\RMSE_\text{LM}$} & \multicolumn{2}{c}{$\RMSE_\text{Volume}$} \\                
        & avg & med &  & avg & med & avg & med & avg & med \\
        \midrule        
        monocular vision   & 5.31 & 4.03 &  & 4.12 & 3.33 & 4.27 & 2.86 & 0.36 & 0.12\\       
        with relative pose & 5.31 & 4.03 &  & 1.75 & 0.63 & 2.39 & 0.89 & 7.28 & 0.13\\        
         \bottomrule
    \end{tabular}
    \caption{Median (med) and average (avg) errors for the different performance metrics used in our evaluation. We can see that the SLAM solution that jointly optimizes the parameters of quadric landmarks and robot poses, corrects the initially estimated map and trajectory layout obtained from odometry measurements.}
    \label{tab:results}
\end{table}

\begin{figure}[t]
    \centering
    \includegraphics[width=0.32\linewidth]{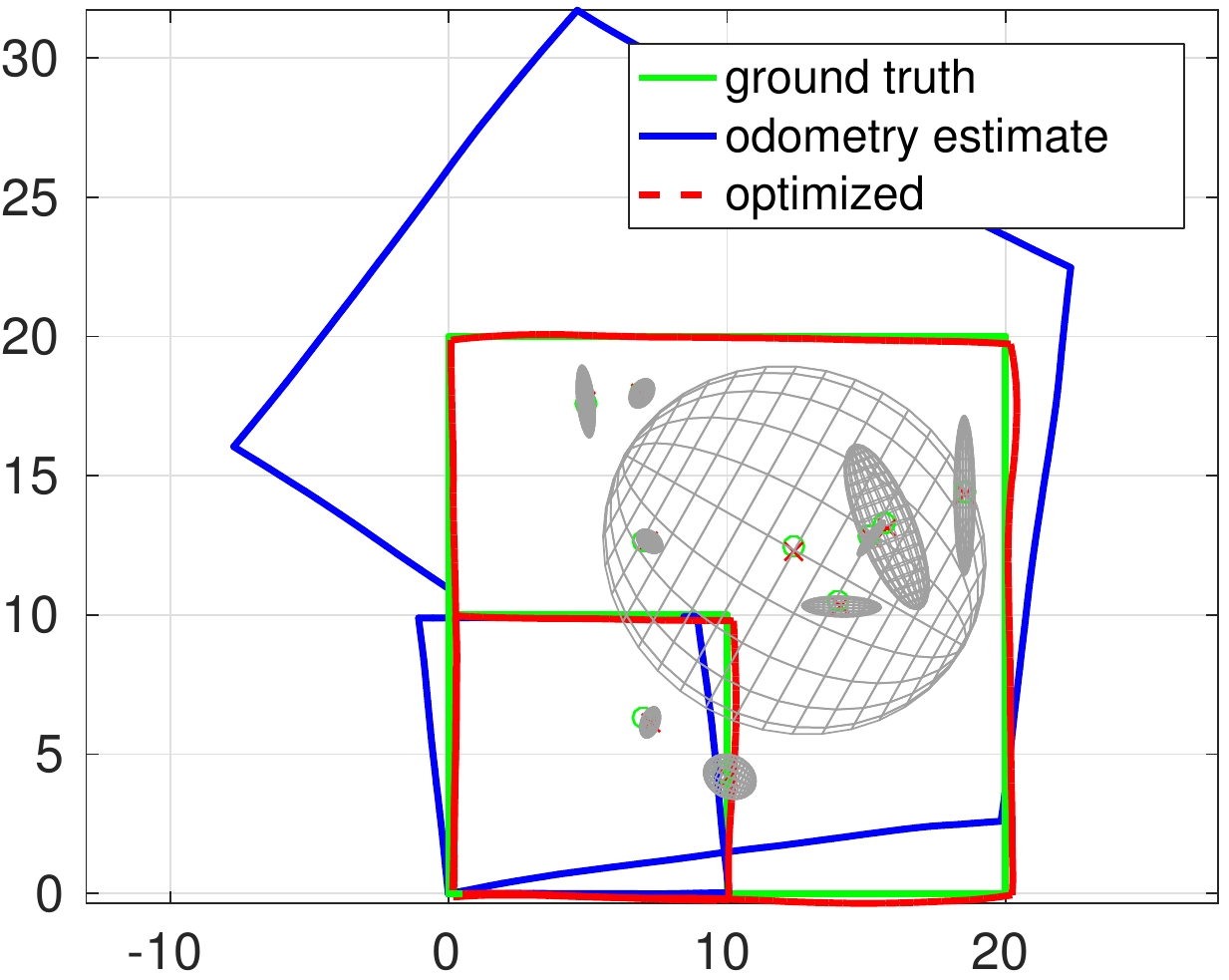}
    \includegraphics[width=0.32\linewidth]{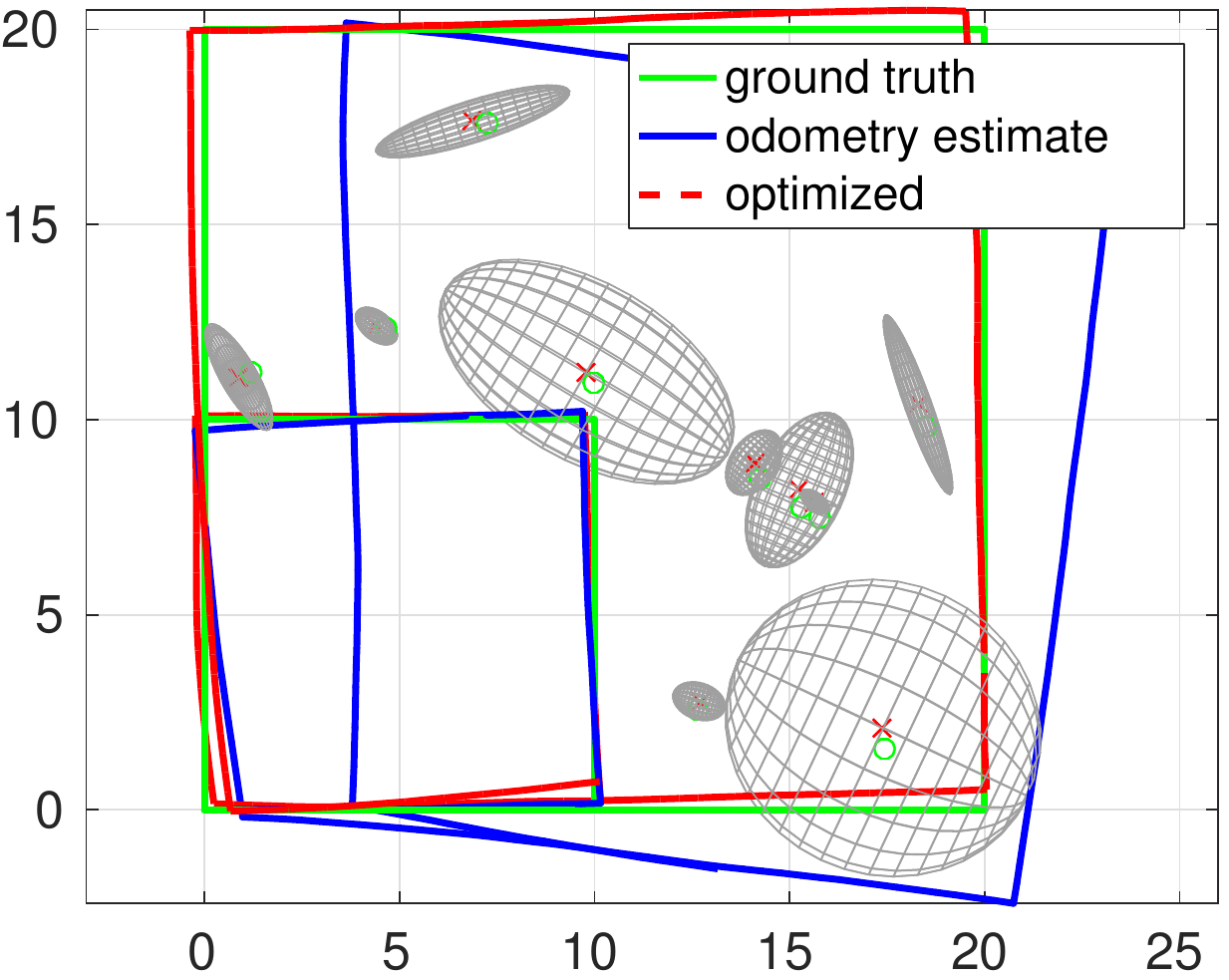}
    \includegraphics[width=0.32\linewidth]{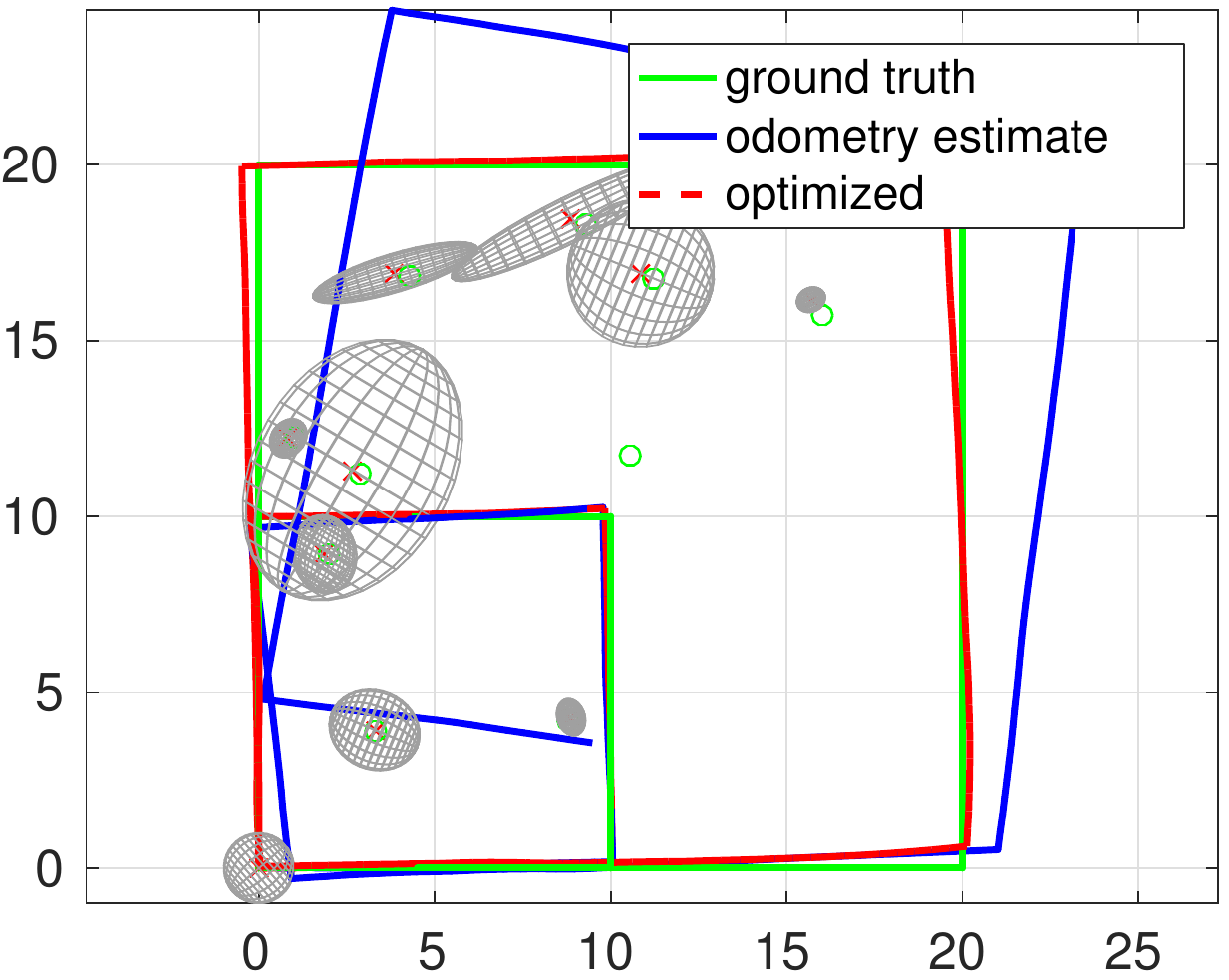}
    \includegraphics[width=0.32\linewidth]{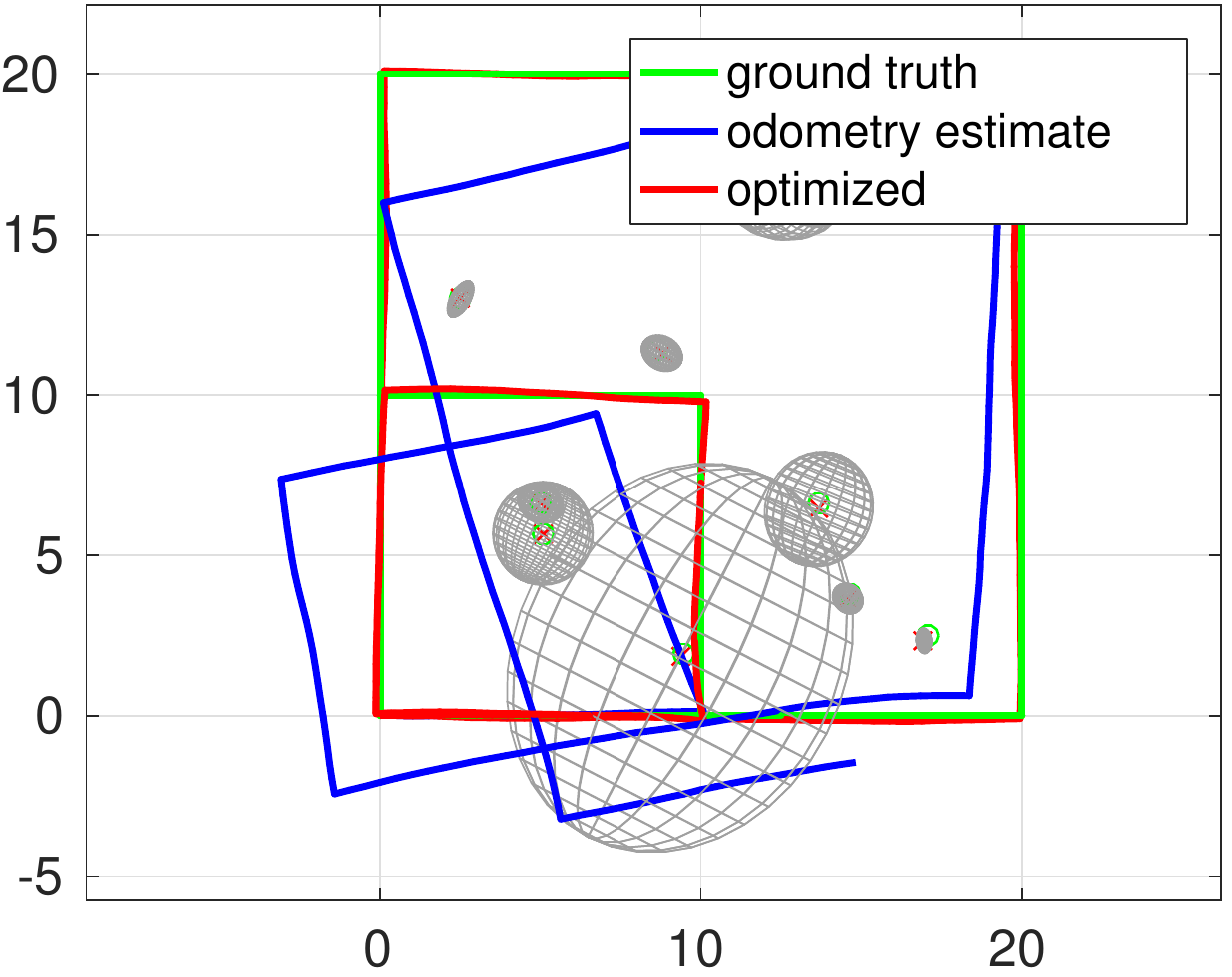}
    \includegraphics[width=0.32\linewidth]{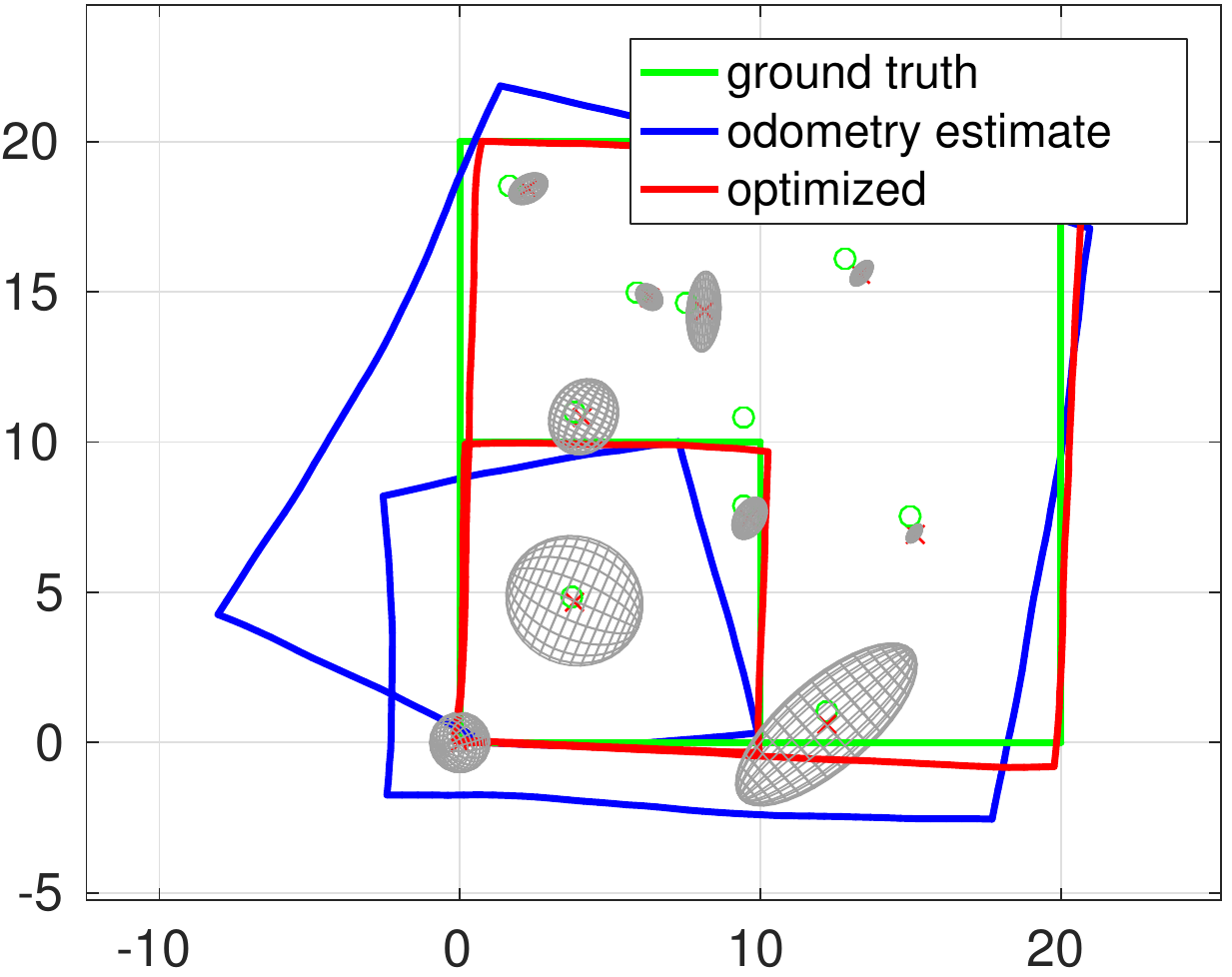}
    \includegraphics[width=0.32\linewidth]{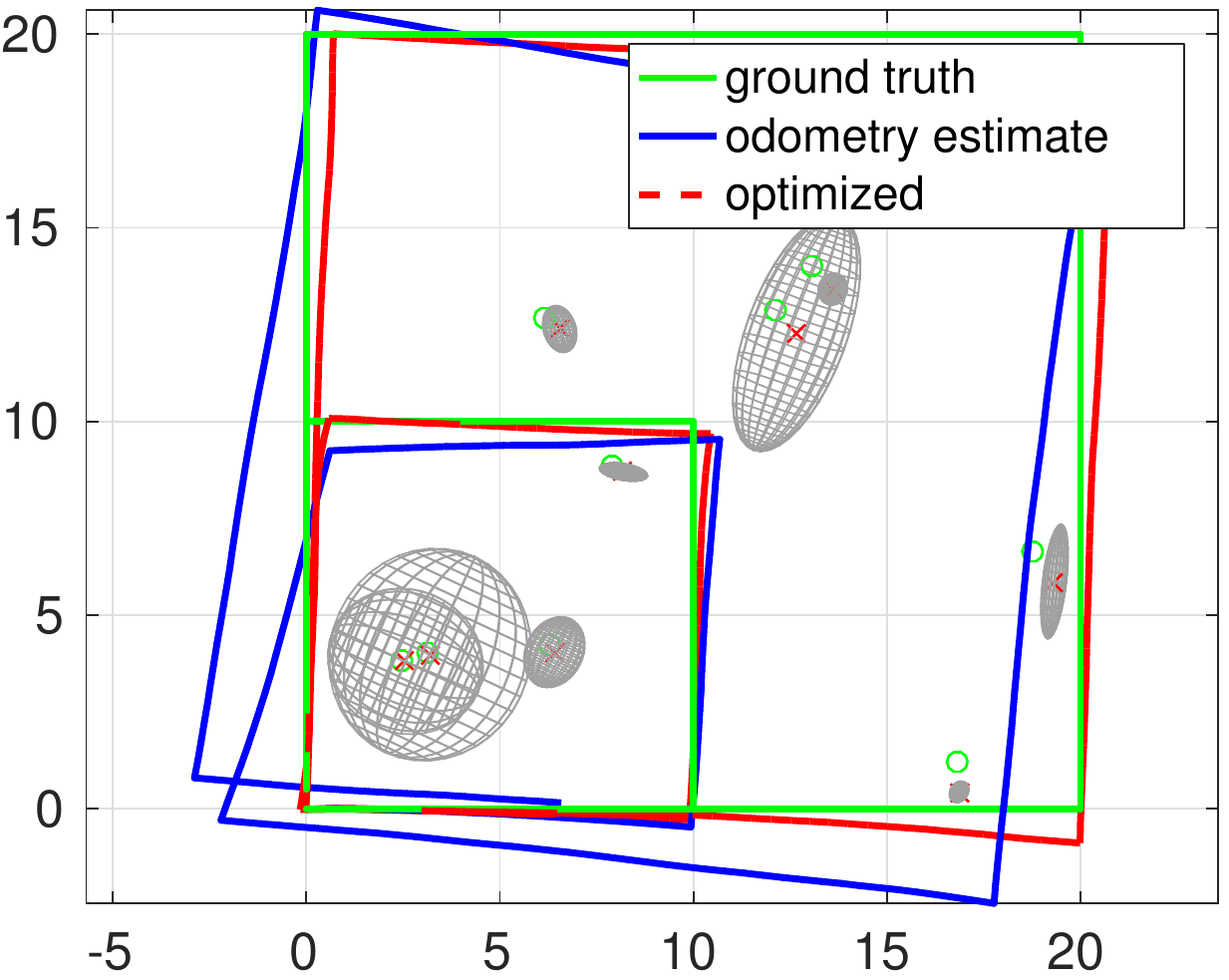}
    \caption{Six example robot trajectories and landmark positions from the evaluation. The initial estimate based on noisy odometry readings is plotted in blue, green shows the ground truth. The SLAM solution obtained by solving the factor graph is illustrated in red and shows how our proposed system can significantly improve both the robot trajectory estimate and the estimated landmark centroids.}
    \label{fig:maps}
\end{figure}

We summarize the results of our experiments in Table \ref{tab:results} and illustrate the initial and optimized maps for randomly chosen trials in Fig. \ref{fig:maps}. We can see that the full SLAM solution (red) significantly improves the quality of the estimated map (robot trajectory and landmark centroids), both in the monocular case and when relative position observations from a depth sensor are available. 

As expected, observing the relative pose of the detected objects improves the quality of the SLAM solution significantly. However, even in the scenario of a monocular camera plus noisy odometry a correcting effect of the quadric landmarks on the the estimated trajectory can be observed, as re-observing the landmarks helps to mitigate the accumulated odometry errors. This loop closure effect is commonly encountered in SLAM and is indeed one of its key characteristics.

The discrepancy between average and median performance indicators hints at the presence of gross outliers in both the estimated robot poses and the landmark quadric parameters. We believe that the strictly planar camera motion leads to a badly constrained system and a difficult optimization problem. This needs to be further assessed in future work.

\section{Conclusions and Future Work}
Our paper demonstrated how to use dual quadrics directly as landmark representations in SLAM with perspective cameras. We derived a factor graph-based SLAM formulation that constrained the dual quadric parameters directly from bounding boxes as they originate from typical object detection systems. The results in this paper demonstrate the utility of such object-based landmarks for SLAM.

Future work will explore how the dual quadrics can be constrained further, e.g. by enforcing an ellipsoid or conic shape. Planar camera movement seems to be a problem case: in our experiments it led to a badly constrained system with unstable convergence properties and local minima in the cost function. Future work should explore these effects in depth and investigate ways to mitigate them. 

We are also working on an efficient implementation of the proposed approach in the C++ based SLAM framework GTSAM. This implementation will enable us to run an evaluation in conjunction with a state of the art object detection system on a real robot.

The advantages of using dual quadrics as landmark parametrizations in SLAM will only increase when incorporating higher order geometric constraints into the SLAM formulation, such as prior knowledge on how landmarks of a certain semantic type can be placed in the environment with respect to other landmarks or general structure. Furthermore, dual quadrics may serve an additional purpose as anchors for more detailed 3D reconstructions of objects. For example, each quadric could carry a local truncated signed distance function that captures the detailed shape information necessary for robotic object manipulation or grasping.

\acknowledgments{This research was conducted by the Australian Research Council Centre of Excellence for Robotic Vision (project number CE140100016). Michael Milford is supported by an Australian Research Council Future Fellowship (FT140101229).}

\bibliography{bibfile}

\end{document}